\documentclass{ecai} 
\usepackage{latexsym}
\usepackage[ruled,noend]{algorithm2e}
\usepackage{amssymb}
\usepackage{amsmath}
\usepackage{amsthm}
\usepackage{booktabs}
\usepackage{enumitem}
\usepackage{enumitem}
\usepackage{environ}
\usepackage{graphicx}
\usepackage{hyperref}
\usepackage{listings}
\usepackage{multirow}
\usepackage{tabularx}
\usepackage{tikz}
\usepackage{stmaryrd}
\usepackage{color}
\usepackage{subcaption}
\usepackage{booktabs}

\usetikzlibrary {shapes.geometric}

\newtheorem{theorem}{Theorem}

\newtheorem{proposition}[theorem]{Proposition}

\newtheorem{definition}{Definition}
\newtheorem{problem}{Problem}
\NewEnviron{problembody}[1][]{%
  \par\noindent%
  \begin{tabularx}{\textwidth}{lX}%
    \BODY\\
  \end{tabularx}%
  \medskip\par%
}
\makeatother

\definecolor{sweetdarkblue}{HTML}{5F9AD4}

\newcommand{\pproblem}{\ensuremath{\Pi}}
\newcommand{\domain}{\mathcal{D}}
\newcommand{\instance}{\mathcal{I}}
\newcommand{\predicates}{\ensuremath{\mathcal{P}}}

\newcommand{\actions}{\mathcal{A}}
\newcommand{\typehierarchy}{\ensuremath{\mathcal{H}}}
\newcommand{\types}{\ensuremath{\mathcal{T}}}
\newcommand{\type}{\ensuremath{\tau}}
\newcommand{\objects}{\ensuremath{\mathcal{O}}}
\newcommand{\init}{\ensuremath{I}}
\newcommand{\goal}{\ensuremath{G}}
\newcommand{\arity}[1]{\ensuremath{\textit{ar}(#1)}}
\newcommand{\issubtype}{\ensuremath{\preceq}}
\newcommand{\pre}[1]{\textsf{pre}\ensuremath{(#1)}}
\newcommand{\add}[1]{\textsf{add}\ensuremath{(#1)}}
\newcommand{\del}[1]{\textsf{del}\ensuremath{(#1)}}

\newcommand{\predvartype}[2]{\ensuremath{\tau}_{#1}(#2)}

\newcommand{\applyop}[2]{\ensuremath{#1[#2]}}

\newcommand{\operator}{\textsf{o}}

\newcommand*{\objclasssymb}{\objects{}}
\newcommand{\env}{\ensuremath{E}}
\newcommand{\envvar}[1]{\ensuremath{\textsf{var}(#1)}}  % Variables an env ranges on

\newcommand{\blozenge}{\ensuremath{\overline{\lozenge}}}
\newcommand{\bsquare}{\ensuremath{\overline{\square}}}
\newcommand{\bbigcirc}{\ensuremath{\overline{\bigcirc}}}
\newcommand{\conset}{\ensuremath{\Lambda}}
\newcommand{\varset}{\ensuremath{\mathcal{X}}}

\newcommand{\tllang}{\ensuremath{\mathcal{L}_{\text{TL}}}}
\newcommand{\ftllang}{\ensuremath{\mathcal{L}_{\text{FTL}}}}

\newcommand{\traces}{\ensuremath{\textit{\textsf{T}}}}
\newcommand{\tscore}{\ensuremath{\sigma}} % Used to be s

\newcommand{\tvalvarsymb}{\ensuremath{y}}  % Truth value: y_i^{t,k}
\newcommand{\tvalenvvar}[4]{\ensuremath{y_{#1}^{#2,#3}[#4]}}  % Truth value with env
\newcommand{\qtfvar}[3]{\ensuremath{\delta_{#1,#2}^{#3}}}  % Quantifier to fluent
\newcommand{\predfvar}[2]{\ensuremath{\theta_{#1}^{#2}}}  % Predicate of a fluent
\newcommand{\logcvar}[2]{\ensuremath{\lambda_{#1}^{#2}}}  % Logical connector of a node
\newcommand{\tsatvar}[1]{\ensuremath{s_{#1}}}  % Satisfaction of the trace by the current formula
\newcommand{\succl}[1]{\ensuremath{\textit{succ}_L(#1)}}
\newcommand{\succr}[1]{\ensuremath{\textit{succ}_R(#1)}}

\newcommand{\exactlyone}[1]{\ensuremath{\textsf{ExactlyOne}_{#1}}}

\newcommand{\BibTeX}{B\kern-.05em{\sc i\kern-.025em b}\kern-.08em\TeX}

\begin{document}

%%%%%%%%%%%%%%%%%%%%%%%%%%%%%%%%%%%%%%%%%%%%%%%%%%%%%%%%%%%%%%%%%%%%%%%%

\begin{frontmatter}

%%% Use this command to specify your submission number.
%%% In doubleblind mode, it will be printed on the first page.

\paperid{918} 

%%% Use this command to specify the title of your paper.

\title{Learning Interpretable Classifiers for PDDL Planning}

%%% Use this combinations of commands to specify all authors of your 
%%% paper. Use \fnms{} and \snm{} to indicate everyone's first names 
%%% and surname. This will help the publisher with indexing the 
%%% proceedings. Please use a reasonable approximation in case your 
%%% name does not neatly split into "first names" and "surname".
%%% Specifying your ORCID digital identifier is optional. 
%%% Use the \thanks{} command to indicate one or more corresponding 
%%% authors and their email address(es). If so desired, you can specify
%%% author contributions using the \footnote{} command.

\author{\fnms{Arnaud}~\snm{Lequen}}

\address{IRIT, Université Toulouse III - Paul Sabatier \\ \emph{Arnaud.Lequen@irit.fr}}

%%% Use this environment to include an abstract of your paper.

\begin{abstract}
We consider the problem of synthesizing interpretable models that recognize the behaviour of an agent compared to other agents, on a whole set of similar planning tasks expressed in PDDL. Our approach consists in learning logical formulas, from a small set of examples that show how an agent solved small planning instances. These formulas are expressed in a version of First-Order Temporal Logic (FTL) tailored to our planning formalism. Such formulas are human-readable, serve as (partial) descriptions of an agent's policy, and generalize to unseen instances. We show that learning such formulas is computationally intractable, as it is an NP-hard problem. As such, we propose to learn these behaviour classifiers through a topology-guided compilation to MaxSAT, which allows us to generate a wide range of different formulas. Experiments show that interesting and accurate formulas can be learned in reasonable time.
\end{abstract}

\end{frontmatter}

%%%%%%%%%%%%%%%%%%%%%%%%%%%%%%%%%%%%%%%%%%%%%%%%%%%%%%%%%%%%%%%%%%%%%%%%

\section{Introduction}

One of the main strengths of PDDL planning models is that they are succinct and human-readable, but can nonetheless express general, complex problems, whose state search spaces are exponential in the size of the encoding \--- as can be the solutions. As a consequence, given a set of examples of the behaviour of an agent (called traces), understanding and recognizing this behaviour can be tedious.% , as the information to process is plentiful. 

In order to summarize the behaviour of a planning agent in a concise, interpretable way, we propose to learn properties that are specific to the solutions proposed by this agent. Such properties, expressed in a temporal logic tailored to fit PDDL planning models, are not only human-readable, but are also general, and can be evaluated against different instances of the same planning problem. This allows them to recognize the behaviour of an agent on instances that are substantially different from the ones used in the set of examples.

More specifically, the problem we tackle is the one where, given a set of positive example traces (the ones of the agent we seek to recognize) and negative examples traces (the ones of other agents), we wish to learn a model that can discriminate as well as possible between positive and negative traces. A wide variety of techniques and models of different natures have been proposed in the literature. Among these, the learning of finite-state automata (DFA) is a well-studied problem~\cite{angluin1987learning,shvo2021interpretable,vaandrager2022new}, but DFAs can grow quickly (thus becoming harder to interpret) and do not generalize to instances not in the example set. More recently, neural network-based architectures such as LSTMs~\cite{zhou2015clstm} have shown very promising results, but lack interpretability, and the rationale for their decision is rarely clear. %In~\cite{hein2018interpretable}, the authors propose to learn sets of algebraic equations through a genetic algorithm, to express a reinforcement learning policy.
% LSTM citation: karim2019multivariate

In the past decade, significant efforts have been made towards learning logical formulas expressed in (a form of) temporal logic. Such works~\cite{raha2022scalable,riener19exact,gaglione2021learning,neider2018learning,camacho2021learning,delarosa2011learning} often leverage symbolic methods to learn Linear Temporal Logic (LTL) formulas~\cite{pnueli1977temporal} that fit the example traces, and thus share some similarities with our work. Some other authors propose other techniques, such as Latent Dirichlet Allocation~\cite{kim2019bayesian}, which stems from the field of natural language processing. 
% However, their method requires that the user gives a pattern on which the algorithm will work: the formulas found by the algorithm will all have the same shape as the pattern. \arnaud{Be more concise with this reference}

% \arnaud{Les idées sont là mais il faut retaper le tout}
However, in all of these cases, the knowledge extracted from the sets of examples has the major drawback of not generalizing well to unknown instances. This is due to the choice of the language used to express these properties. For instance, since LTL formulas are built over a set of propositional variables, they do not generalize to models that do not share the same variables.

To address this issue, we propose to learn properties in a version of First-Order Temporal Logic (FTL). When tailored to the PDDL planning formalism, FTL can express a wide range of properties that generalize from one planning instance to the other, given that they model similar problems. This was shown in~\cite{bacchus00using}, who proposed to express \emph{search control knowledge} in a language similar to ours, albeit with the aim of guiding the search of a planner designed to use such knowledge. In~\cite{delarosa2011learning}, the authors proposed to synthesize such control knowledge automatically, and thus address the problem of learning properties expressed in a fragment of FTL.

In this paper, we show that it is possible to learn richer and more expressive properties, using the whole range of FTL operators and modalities. The properties we wish to learn should describe the behaviour of a given planning agent, without being true for the behaviour of other agents. We show that learning such formulas is computationally intractable, as the associated decision problem is NP-hard. This is why the core of our approach consists in encoding the learning problem into a MaxSAT instance, which has the added benefit of showing resilience to any potential noise in the set of training examples. To make the search more efficient, we fix the general topology of the target formula before the encoding. In addition to alleviating the load on the MaxSAT solver and rendering the algorithm more parallelizable, this also increases the diversity in the formulas learned by our algorithm, thus providing varied descriptions of the behaviour of the agent of interest.

Our article is organised as follows: Section~\ref{sec:background} introduces the planning formalism as well as the FTL language. Section~\ref{sec:ftllearningproblem} formally introduces the learning problem we tackle in this paper, and shows that the associated decision problem is intractable. Sections~\ref{sec:preprocessing} and \ref{sec:topologyguiding} present some technical choices that we made to solve our problem in reasonable time in practice.
In Section~\ref{sec:encoding}, we describe our reduction of the problem to MaxSAT, and in Section~\ref{sec:experiments}, we present our experimental results, as well as a few examples of formulas that are within reach of our implementation.

\section{Background}
\label{sec:background}

\subsection{Planning with PDDL}

This section introduces the model that we use to describe planning tasks. Our definition of a PDDL planning task differs from~\cite{helmert2009concise}, as we require the organization of the objects of our instances into types. The model we use resembles the one defined in~\cite{horvcik2021endomorphisms}

\begin{definition}[Type tree]
    A \emph{type tree} $\mathcal{T}$ is a non-empty tree where each node is labeled by a symbol, called a \emph{type}. For any type $\type \in \mathcal{T}$, we call \emph{strict subtype} any descendant $\type'$ of $\type$. $\type'$ is a subtype of $\type$ (denoted $\type' \issubtype{} \type$) when $\type'$ is a strict subtype of $\type$ or when $\type' = \type$.
\end{definition}

\begin{definition}[Object class]
Let \objects{} be a set of elements called \emph{objects}. We call \emph{object class} any subset of \objects{}. A class $c_i$ is said to be a \emph{subclass} of type $c_j$ if $c_i \subseteq c_j$. 
\end{definition}

% In the following, for every object $o \in \objects{}$, we will denote $\vartype{o} \in \types{}$ the smallest type of $\types$ (in the sense of inclusion) containing object $o$.

\begin{definition}[Type hierarchy]
    A \emph{type hierarchy} $\typehierarchy$ over type tree $\types$ is a set of object classes such that $\objects{} \in \typehierarchy$, and such that each object class of $\typehierarchy$ is mapped to a unique type of $\types$. This mapping $\type: \typehierarchy \rightarrow \types $ is such that for any pair $c_i, c_j$ of object classes:
    \begin{itemize}
        \item $c_i$ is a subclass of $c_j$ iff $\type(c_i)$ is a subtype of $\type(c_j)$;
        % \item $c_j$ is a subclass of $c_i$ iff $\type(c_j)$ is a subtype of $\type(c_i)$;
        \item $c_i \cap c_j = \varnothing$ iff $\type(c_i)$ is not a subtype of $\type(c_j)$ (and conversely).
    \end{itemize}
    We say that object $o \in \objects{}$ is of type $\type{}(o) := \type{}(c)$ where $c$ is the smallest (for inclusion $\subseteq$) class of $\typehierarchy$ to which $o$ belongs.
    % and for each pair $\tau_i, \tau_j \in \types$, either $\tau_i$ is a subtype of $\tau_j$, or $\tau_j$ is a subtype of $\tau_i$, or $\tau_i \cap \tau_j = \varnothing$.
\end{definition}

\begin{definition}[Predicate, atoms and fluents]
    A \emph{predicate} $p$ is a symbol, with which is associated:
    \begin{itemize}
        \item An arity $\arity{p} \in \mathbb{N}$
        \item A type for each of its arguments. For $i \in \{1, \ldots, \arity{p}\}$, the type of its argument at position $i$ is denoted $\predvartype{p}{i} \in \types{}$
    \end{itemize}
    
    An \emph{atom} is a predicate for which each argument is associated with a symbol, which can be a variable symbol, or an object of $\objects{}$. When the $i$-th argument of the atom is an object $o \in \objects{}$ (associated to type hierarchy $\typehierarchy$), then we require that $\type{}(o) = \predvartype{p}{i}$.
    The atom consisting of predicate $p$ and symbols $x_1, \ldots, x_{\arity{p}}$ is denoted $p(x_1, \ldots, x_{\arity{p}})$.
    %, \arnaud{Remove this?} or simply $p(\vec{x})$ with the shorthand $\vec{x} = (x_1, \ldots, x_{\arity{p}})$.

    A \emph{fluent} is an atom where each argument corresponds to an object of $\objects{}$. A \emph{state} is a set of fluents.
\end{definition}

\begin{definition}[Action schema and operators]
    An \emph{action schema} is a tuple $a = \langle \pre{a}, \add{a}, \del{a} \rangle$, such that $\pre{a}$, $\add{a}$ and $\del{a}$ are sets of \emph{atoms} instantiated with variables only.
    % on which $\params{a}$ is defined, and
    % \begin{itemize}
    %     \item $\params{a}$ is a sequence of variable symbols $x$, which have a type $\vartype{x} \in \types$.
    %     \item $\pre{a}$, $\add{a}$ and $\del{a}$ are sets of \emph{atoms}, which are predicates instantiated with variables $x \in \params{a}$.
    % \end{itemize}

    An \emph{operator} $\textsf{o}$ is akin to an action schema, except that the sets $\pre{\textsf{o}}$, $\add{\textsf{o}}$ and $\del{\textsf{o}}$ are sets of fluents.
\end{definition}
%
% This leads us to the main definition of this section, which crystallizes the elements above.
%
\begin{definition}[PDDL planning problem]
    A PDDL planning problem is a pair $\pproblem{} = \langle \domain{}, \instance{} \rangle$ where
    $\domain = \langle \predicates{}, \actions{}, \types{} \rangle$ is the domain and $\instance{} = \langle \objects{}, \typehierarchy, \init, \goal \rangle$ is the instance.
    
    The domain $\domain$ consists of a set $\predicates{}$ of predicates, a set of actions schemas $\actions{}$, and a type hierarchy $\types$.
    
    The instance $\instance{}$ consists of a set of objects $\objects{}$ and an associated type hierarchy $\typehierarchy$, as well as two states, $\init$ and $\goal$, which are respectively the initial state and the goal conditions.
\end{definition}

An operator $\operator{}$ is applicable in a state $s$ if $\pre{\operator{}} \subseteq s$. The state that results from the application of $\operator{}$ in $s$ is $\applyop{s}{\operator{}} = \left( s \setminus \del{\operator{}} \right) \cup \add{\operator{}}$.

A sequence of operators $\operator{}_1, \ldots, \operator{}_n$ is called a \emph{plan} for $\pproblem{}$ if there exists a sequence of states $s_0, \ldots, s_n$ where $s_0 = \init$, and which is such that, for all $i \in \{1, \ldots, n\}$, $s_i = \applyop{s_{i-1}}{\operator{}_i}$ and $\operator{}_i$ is applicable in $s_{i-1}$. Such a sequence of states (which is unique for each plan) is called a \emph{trace}. A plan is called a \emph{solution-plan} if, in addition to this, $\goal \subseteq s_n$.
We say that a fluent $p(o_1, \ldots, o_{\arity{p}})$ is true in state $s$ iff $p(o_1, \ldots, o_{\arity{p}}) \in s$.

\subsection{First-Order Temporal Logic (FTL)}
\label{sec:ftllang}

% This section introduces the language in which are expressed the formulas that we attempt to learn throughout this paper. 

\paragraph{Syntax}
Let \varset{} be a set of variable symbols, \predicates{} a set of predicates, and \types{} 
a type tree. %built over the set of objects \objects{}.
We define our language $\ftllang{}$ such that:
\begin{align*}
    \psi :=
    \exists x \in \type. \psi \mid 
    \forall x \in \type. \psi \mid 
    \varphi    
\end{align*}
where $\varphi \in \tllang$, and $\tllang$ is such that:
\begin{align*}
\varphi := \; &\top \mid
p(x_1, \ldots, x_{\arity{p}}) \mid 
\neg \varphi \mid
\bigcirc \varphi \mid
\lozenge \varphi \mid 
\square \varphi \mid
\bbigcirc \varphi \mid
\blozenge \varphi \mid
\bsquare \varphi \mid \\
&\varphi \,\textsf{U}\, \varphi \mid
\varphi \wedge \varphi \mid
\varphi \vee \varphi \mid
\varphi \Rightarrow \varphi
\end{align*}
where $x_1, \ldots, x_{\arity{p}}$ are variables of \varset{}, $p$ a predicate of \predicates{}, and $\type$ a type.
In the following, we will denote $\conset{} = \{\wedge, \vee, \Rightarrow, \textsf{U}, \bigcirc, \lozenge, \square\}$ the set of all logical operators. For each operator $\lambda \in \conset{}$, we also note $\arity{\lambda} \in \{1, 2\}$ the arity of the operator.

This formulation is akin to Linear Temporal Logic on finite traces (LTL$_\text{f}$)~\cite{pnueli1977temporal}, where propositional variables are replaced with first-order predicates and variables. Notice that we only work with formulas in prenex normal form.

\paragraph{Environments}
The formulas of $\tllang$ are built on atoms whose arguments are variables of $\varset{}$, while traces contain fluents. We bridge that gap with the notion of \emph{environments}, which are akin to interpretations in first-order logic.
% As we work with a logic reminiscent of first-order logic (on finite domains), we are required to reason on various sets of objects at once. The concept of environment allows us to work on assignations of (sets of) objects to variables of the formula.

Let us denote $\varset{} = (x_1, \ldots, x_q)$. In addition, let $\instance$ be an instance, with objects $\objects = \{o_1, \ldots, o_{\vert \objects \vert}\}$. We call a \emph{partial} environment any assignment of some of the variables $x_1, \ldots, x_q$ to objects of $\objects$. Let us denote $\envvar{e}$ the variables that are assigned an object within the partial environment $e$. When $\envvar{e} = \varset{}$, we simply say that $e$ is an environment.
We denote any (partial) environment $e = \{x_1 := o_{i_1}, \ldots, x_q := o_{i_q}\}$, where $i_1, \ldots, i_q \in \llbracket 1, \vert \objects \vert \rrbracket$.

The object to which variable $x$ is associated to in $e$ is denoted $x[e]$. We also denote $p(x, y)[e]$ the grounding of an atom $p(x, y)$ by an environment $e$ such that $x, y \in \envvar{e}$. If $e = \{x := o_1, y := o_2, \ldots\}$, then $p(x, y)[e] = p(x[e], y[e]) = p(o_1, o_2)$. By extension, the formula obtained when grounding each atom of $\varphi$ with $e$ is written $\varphi[e]$.

\paragraph{Semantics}
Given an environment $e$, any quantifier-free formula $\varphi$ of $\tllang$ can be evaluated against a trace $t = \langle s_0, \ldots, s_n \rangle$, at any step. When $i \in \llbracket 0, n \rrbracket$, we write $t, e, i \models \varphi$ to denote that formula $\varphi$ is true at state $s_i$ of trace $t$ with environment $e$. Temporal modalities, such as $\bigcirc$, $\lozenge$, $\square$, etc., are used to reason over the states that follow or precede the current state $s_i$.

$\bigcirc \varphi$ means that property $\varphi$ is true in the next state, while $\lozenge \varphi$ means that $\varphi$ is eventually true, in one of the successors of the current state. $\square \varphi$ means that $\varphi$ is true from this state on, until the end of the trace, and $\varphi_1 \textsf{U} \varphi_2$ means that $\varphi_2$ is true in some successor state, and until then, $\varphi_1$ is true. Operators $\bbigcirc{}$, $\blozenge{}$ and $\bsquare{}$ are the \emph{past} counterparts of the previous connectors: $\bbigcirc{} \varphi$ means that $\varphi$ is true in the previous state, $\blozenge \varphi$ that $\varphi$ is true in some previous state, and $\bsquare \varphi$ that $\varphi$ is true in every previous state.

To illustrate the language, we introduce the Childsnack problem, which originates from the International Planning Competition (IPC). It consists in making sandwiches and serving them to a group of children, some of whom are allergic to gluten. Sandwiches can only be prepared in the kitchen, and then have to be put on trays, which is the only way they can be brought to the children for service. Among the following FTL formulas, the first indicates that ``All children will eventually be served'' (and will be satisfied by any solution-plan). The second formula indicates that every sandwich $x$ will eventually be put on some tray, at a moment $t+1$. For every moment that precedes moment $t$, $x$ will not be prepared yet (which indicates that the sandwich is actually put on the tray right after being prepared).
\begin{align}
    \forall x \in \text{Child}. \, \lozenge \, \text{served}(x) \\
    \forall x \in \text{Sandwich}.\, \exists y \in \text{Tray}.\,  \text{notprepared}(x)\, \textsf{U}\, \bigcirc \text{on}(x, y) \label{eq:notprepareduntillastminute}
\end{align}
%
%Note that, since the formulas of $\ftllang$ do not contain constants or objects from a planning instance, they can be evaluated against any instance.

Temporal modalities can be expressed in terms of one another.
%\-- the same way propositional connectors can be expressed in terms of one another.
For any quantifier-free formula $\varphi$, we have $\lozenge \varphi \equiv \top \textsf{U} \varphi$, $\square \varphi \equiv \neg \lozenge \neg \varphi$ and $\bsquare \varphi \equiv \neg \blozenge \neg \varphi$. This leads us to an inductive definition of the semantics of our language, for quantifier-free formulas of $\tllang$:
\begin{align*}
    &t, e, i \models p(x, \ldots, x) &\text{iff }& p(x, \ldots, x)[e] \in s_i \\
    &t, e, i \models \neg \varphi &\text{iff }& t, e, i \not\models \varphi \\
    &t, e, i \models \varphi_1 \wedge \varphi_2 &\text{iff }& t, e, i \models \varphi_1 \text{ and } t, e, i \models \varphi_2 \\
    &t, e, i \models \bigcirc \varphi &\text{iff }& i < n \text{ and } t, e, (i+1) \models \varphi \\
    &t, e, i \models \bbigcirc \varphi &\text{iff }& i > 0 \text{ and } t, e, (i-1) \models \varphi \\
    &t, e, i \models \blozenge \varphi &\text{iff }& \exists j \in \llbracket 0, i\rrbracket \text{ s.t. } t, e, j \models \varphi \\
    &t, e, i \models \varphi_1 \textsf{U} \varphi_2 &\text{iff }& \exists j \in \llbracket i, n\rrbracket \text{ s.t. } t, e, j \models \varphi_2 \\
    & & &\text{and } \forall k \in \llbracket i, j-1\rrbracket, t, e, k \models \varphi_1
\end{align*}

We write $t, e \models \varphi$ as a shorthand for $t, e, 0 \models \varphi$, which means that trace $t$ satisfies the formula $\varphi$, since it is true in the initial state of $t$.
A formula $\psi \in \ftllang$ is evaluated against instantiated traces:%, as defined below: %against pairs $\langle t, \instance \rangle$, where $t$ is a trace and $\instance$ is an instance associated to some domain $\domain$. 
\begin{definition}[Instantiated trace]
    An \emph{instantiated trace} is a pair $\langle t, \instance\rangle$ such that $t$ is a trace where fluents are built on the objects \objects{} of the planning instance $\instance$.
    %\arnaud{Make the link between traces and instances clearer, and earlier}
\end{definition}

%For any formula $\phi$ of $\ftllang$, let us denote $\phi \varreplace{x}{y}$ the formula of $\ftllang$ where each occurrence of $y$ is replaced by $x$ \arnaud{in the $\tllang$ formula}. 

For any partial environment $e$, $x \in \varset$, $o \in \objects$, we denote $e[x := o]$ the environment identical to $e$, but where variable $x$ is associated $o$.
The semantics of $\ftllang$ is defined as follows:
\begin{align*}
    &\langle t, \instance \rangle, e \models \forall x \in \type. \psi &\text{iff }& \text{for all } o \in \objects \text{ s.t. } \type{}(o) = \type, \\
    & & & \langle t, \instance \rangle, e[x := o] \models \psi \\
    &\langle t, \instance \rangle, e \models \exists x \in \type. \psi &\text{iff }& \text{there exists } o \in \objects \text{ s.t. } \type{}(o) = \type, \\
    & & & \langle t, \instance \rangle, e[x := o] \models \psi \\
    &\langle t, \instance \rangle, e \models \varphi &\text{iff }& t, e \models \varphi
\end{align*}
%
% \begin{align*}
%     &\langle t, \instance \rangle \models \forall x \in \type. \psi &\text{iff }& \text{for all } o \in \objects \text{ s.t. } \type{}(o) = \type, \\
%     & & & \langle t, \instance \rangle \models \psi\varreplace{o}{x} \\
%     &\langle t, \instance \rangle \models \exists x \in \type. \psi &\text{iff }& \text{there exists } o \in \objects \text{ s.t. } \type{}(o) = \type, \\
%     & & & \langle t, \instance \rangle \models \psi\varreplace{o}{x} \\
%     &\langle t, \instance \rangle \models \varphi &\text{iff }& t \models \varphi
% \end{align*}
%
where $x$ is a variable, and $\varphi$ is a formula of $\tllang$ (thus quantifier-free). We will often denote $\langle t, \instance \rangle \models \psi$ as a shorthand for $\langle t, \instance\rangle, \varnothing \models \psi$, where $\varnothing$ is the empty environment.

Note that it is well known that the past modalities do not change the expressivity of LTL. As a consequence, our language could have expressed the same properties without modalities $\bbigcirc$, $\blozenge$ or $\bsquare$. However, we include these modalities in our language as they may make some properties more succinct to express~\cite{markey2003temporal}.

% \arnaud{We should give an example of a formula. For instance, we could give a formula that describes the goal of childsnack: $\forall x \in Children. \lozenge \text{served}(x)$. Say that this can express all sorts of things, and in this paper, we use it to express something that describes the behaviour of an agent}

% \paragraph{Planning-related modalities}
% \arnaud{Introduce this later, because otherwise it's a bit annoying, there are lost of things to change (eg. we have to add the instance to the model of every formula, even quantifer-free)}
% Maybe allow a goal modality (for instance $\textsf{G}$) that dictates that the fluent under that modality is in the goal?

\subsection{The MaxSAT problem}

Let Var be a set of propositional variables. The boolean satisfiability problem (SAT) is concerned with finding a valuation that satisfies a propositional formula $\phi$. Propositional formulas are defined as follows, where $x \in \text{Var}$ is a propositional variable:
\[
    \phi := \top \mid x \mid \neg \phi \mid \phi \vee \phi \mid \phi \wedge \phi
\]
The maximum boolean satisfiability problem (MaxSAT) is a variant of SAT, in which a valuation of the variables Var of a set of formulas $\{\phi_1, \ldots, \phi_n\}$ is sought. Each formula $\phi_i$ is assigned a \emph{weight} $w(\phi_i) \in \mathbb{R} \cup \{\infty\}$. The MaxSAT problem consists in finding a valuation $v$ of Var such that the sum of the weights of the formulas that are not satisfied by $v$ is minimal.

\section{The $\ftllang$ learning problem}
\label{sec:ftllearningproblem}

\subsection{Problem definition}

% In this section, we introduce the main problem we tackle in this paper. 

\paragraph{Score function}

Our problem takes in input a \emph{score function}, denoted $\tscore: \traces \rightarrow \mathbb{R}$, where $\traces$ is the set of traces. This function allows us to express preferences on which traces are the most important to capture in the output formula, and which traces are the most important to avoid. In the rest of this article, we will say that an instantiated trace $\langle t, \instance \rangle$ is \emph{positive} iff $\tscore{}(\langle t, \instance\rangle) \geq 0$. Otherwise, the instantiated trace is said to be \emph{negative}. 

In the following, we use $\left[ \langle t, \instance \rangle \models \psi \right]$ as a shorthand for the function equal to $1$ if $\langle t, \instance \rangle \models \psi$ and equal to $0$ otherwise. The score function generalizes to formulas $\psi \in \ftllang$ as follows:
\[ 
    \tscore{}^{\traces}(\psi) = \sum_{\langle t, \instance{} \rangle \in \traces} \hspace{1ex} \tscore(\langle t, \instance \rangle)\left[ \langle t, \instance \rangle \models \psi \right]
\]

\begin{problem}{$\ftllang$ learning}
    \label{prob:foltlf_learning}
    \begin{problembody}
        \emph{\textbf{Input}}: & $\domain$ a domain\\
                               & $\traces$ a set of instantiated traces \\
                               & $r \in \mathbb{N}$ the maximum number of logical \\
                               & \hspace{3ex} operators in the output formula \\
                               & $q \in \mathbb{N}$ the maximum number of quantifiers \\
                               & $\tscore: \traces \rightarrow \mathbb{R}$ a function called the \emph{score function}\\
        \emph{\textbf{Output}}: & A formula $\psi \in \ftllang$ such that $\psi$ has at most \\
        & $r$ logical operators, and $q$ quantifiers, and \\
        & \hspace{4ex} $\sum_{\langle t, \instance{} \rangle \in \traces} \hspace{1ex} \tscore(\langle t, \instance \rangle)\left[ \langle t, \instance \rangle \models \psi \right]$ \\
        & is maximal
    \end{problembody}
\end{problem}
%
% In the rest of this paper, we will use the notations introduced in the input of Problem~\ref{prob:foltlf_learning}. In addition, we will denote $\traces := \postraces \cup \negtraces$ the set of all traces, positive or negatives.
%
\subsection{Complexity}

% \subsubsection{Hardness bound}
The decision problem associated to the $\ftllang$ learning problem is the problem for which the output is \emph{Yes} iff \emph{there exists a formula $\psi \in \ftllang$ satisfying the requirements above, and with score $\tscore{}^\traces(\psi) \geq \ell$}, where $\ell$ is given in input. The proof of intractability consists in a reduction from the NP-hard Set Cover Problem~\cite{karp1972reducibility}. It is sketched below, and a more detailed proof can be read in the supplementary material of this article~\cite{lequen2024proof}.

\begin{proposition}
    \label{prop:np_hardness}
    The decision problem associated to the $\ftllang$ learning problem is NP-hard
\end{proposition}

%
%
% WHEN WRITING THE THESIS, THE FOLLOWING CAN BE ADDED
%
%

% The main difficulty that rises when working with environments is that there are $\vert \objects \vert^q$ different environments. As $\objects$ can be large, the number of environments being exponential in the number of quantifiers quickly makes the problem intractable, if no restriction is posed. This is why we ensure that each variable of the \ftllang{} formula to learn is assigned a type, so that not every environment has to be considered during search: the types are chosen before proceeding to the encoding.
% which drastically restricts the number of possibilities considered. The object classes are chosen and affected before we proceed to the encoding.

% We now introduce the problem we reduce to our problem:

\begin{problem}{Set Cover}
    \label{prob:set_cover}
    \begin{problembody}
        \emph{\textbf{Input}}: & A set $U = \{1, \ldots, n\}$\\
                               & A set $S$ of subsets: $S = \{S_1, \ldots, S_m \} \subseteq 2^U$ \\
                               & $k \in \mathbb{N}$\\
        \emph{\textbf{Output}}: & Yes \emph{iff} there exists a subset $T \subseteq S$ such that $\vert T \vert \leq k$ \\ & and $\cup_{s \in T} s = U$ \\
        & No \emph{otherwise}
    \end{problembody}
\end{problem}

\begin{proof}[Proof of Proposition~\ref{prop:np_hardness} (Sketch)]
    Let us consider an instance of \emph{Set Cover}. We build an instance of \emph{$\ftllang$ learning} that is positive (i.e. outputs \emph{Yes}) \emph{iff} the \emph{Set Cover} instance is positive.

    The proof consists in showing that a set of positive traces can be described by a formula $\ftllang$ satisfying the constraints in input iff there exists a set cover of size at most $k$. Each of the positive traces is associated to an element $j$ of $U$, and contains a single state (and thus, temporal modalities have no effect). This single state carries the information as to which subsets contain $j$. The information consists of fluents of the form $\text{in}(S_i)$. For instance, if only sets $S_1, S_3 \in S$ contain element $j$, then the $j$-th trace is $\langle \{\text{in}(S_1), \text{in}(S_3)\} \rangle$.
    
    Each subset $S_i$ of $S$, in the Set Cover instance given in input, is associated a unique type $\text{Set}_i$ in the PDDL domain we build. As our $\ftllang$ formulas quantify on these types, and since we restrict the number of quantifiers to $q = k$, any output formula of the $\ftllang{}$ learning problem can only reason on at most $k$ subsets among those in $S$. If $k$ quantifiers are enough to distinguish between the positive traces (associated to elements $j \in U$) and a mock trace, then $k$ subsets are enough to cover all elements of $U$. Otherwise, no set cover of size at most $k$ exists.
    
    % If we can learn a formula that distinguishes between all the positive traces and a mock trace, using at most $q = k$ quantifications, then there exists a set cover of $U$ that uses at most $k$ subsets. Otherwise, no set cover exists.
    
    More formally, let us define the input of our \emph{$\ftllang$ learning} instance as follows:
    the domain is $\domain = \langle \predicates{}, \actions{}, \types{} \rangle$, where types are $\types = \{ \text{Set}_1, \ldots, \text{Set}_m, \text{Set} \}$,
    $\predicates{} = \{\text{in}\}$ (with $\arity{\text{in}} = 1$ and $\predvartype{\text{in}}{1} = \text{Set}$), and $\actions{} = \varnothing$ (as all traces have length 1, no action is needed).

    Each instantiated trace is associated its own instance, which is in turn associated to a unique element $j \in U$: for $j \leq n$, $\instance{}_j = \langle \objects{}, \typehierarchy, \init_j, \goal_j \rangle$. The shared set of objects contains the sets of the Set Cover input instance, and is such that $\objects{} = \{S_1, \ldots, S_m, d\}$ (where $d$ is a mock object that we use to discriminate traces later). The type hierarchy $\typehierarchy{} = \{\{ S_i \} \mid S_i \in S \} \cup \objects$ associates each set to its own unique type in our PDDL domain: types are such that $\type{}(S_i) = \text{Set}_i$ for all $i \leq m$, and $\type{}(\objects{}) = \text{Set}$.
    %Each element $j \in U$ is associated a unique planning instance $\instance_j$. 
    The information regarding which sets $j$ belongs to is encoded in the initial state and goal of $\instance_j$: for $j \leq n$, $\init{}_j = \goal_j = \{\text{in}(S_i) \mid j \in S_i\}$.
    
    We also define a mock, negative instance $\instance_d = \langle \objects, \typehierarchy, \init_d, \goal_d \rangle$, with $\init_d = \{\text{in}(d)\}$, and $\goal_d = \init_d$. We have $\traces = \{ \langle t_j, \instance_j \rangle \mid j \leq n \} \cup \{\langle t_d, \instance_d \rangle\}$, where $t_j = \langle I_j \rangle$ and $t_d = \langle I_d \rangle$. We set $r = m - 1$, $q = k$, and the score $\tscore{}(\langle t_j, \instance_j\rangle)$ of every trace to 1, except $\langle t_d, \instance_d \rangle$ for which $\tscore{}(\langle t_d, \instance_d \rangle) = -1$. The decision problem consists in finding whether there exists a formula $\psi \in \ftllang$ that is satisfied by every trace, except $\langle t_d, \instance_d \rangle$. Thus, the score threshold we set is $\ell = m$.
    Let us now show that this instance is positive iff our \emph{Set Cover} instance is positive. 
    
    Suppose that there exists a set cover $T = \{S_{i_1}, \ldots, S_{i_k}\}$. Then there exists a formula $\psi \in \ftllang{}$ with score exactly $m$:
    \[
        \psi := \exists x_1 \in \text{Set}_{i_1}, \ldots, \exists x_k \in \text{Set}_{i_k}.
        \bigvee_{j \leq k} \text{in}(x_j)
    \]
    We have that $\langle t_d, \instance_d \rangle \not\models \psi$, but all other traces satisfy $\psi$, since $T$ is a set cover. So the formula has score exactly $m$, and is a positive instance of \emph{$\ftllang$ learning}.
    
    Now suppose that there exists no set cover of size at most $k$ for the input instance. By contradiction, suppose that there exists a formula $\psi$ with score $m$. % (which is the maximum possible score).
    We denote $T'$ the set of types on which $\psi$ quantifies upon. We have that $\vert T' \vert \leq k$.
    Since there exists no set cover of size at most $k$, then there exists $w \leq n$ such that $w \not\in \cup_{s \in T'} s$. The rest of the proof consists in showing that if $\langle t_w, \instance_w \rangle \models \psi$, then $\langle t_d, \instance_d \rangle \models \psi$, which contradicts the fact that $\psi$ has score $m$. This can be proven by structural induction on $\psi$.

    Since the $\ftllang$ learning instance we built is positive iff the set cover instance is positive, the $\ftllang$ learning problem is NP-hard.
\end{proof}

In~\cite{fijalkow21complexity, mascle2023learning}, the authors tackle the $\mathcal{L}_{\text{LTL}}$ learning problem, where $\mathcal{L}_{\text{LTL}}$ is the fragment of $\tllang$ without past modalities. Along with other past modalities-free fragments, they showed the corresponding decision problem to be NP-complete. 
% \arnaud{STOPPED HERE : they showed it to be ..., but our proof is better suited for our case because...}

% Various authors tried to settle the complexity of problems related to ours, without always succeeding, even when dealing with simpler languages. Notable works include~\cite{fijalkow21complexity}, where the authors show interest in the problem of learning various fragments of LTL. Even though the learning problems associated to several fragments were shown to be NP-complete, the complexity of the problem associated to the whole language is still open. \arnaud{Rewrite this. Say that the proof where there are no quantifier but LTL connectors is out there}

% For the sake of clarity, we will often apply the score function directly to the trace, without specifying the instance of the instantiated trace.

% Various fragments of LTL were showed to be hard for the associated learning problem~\cite{fijalkow21complexity} (which is however simpler than our case, as we also consider first-order quantifications)

Given an environment $e$, a trace $t$, and a formula $\varphi \in \tllang$, checking that $t, e \models \varphi$ can be done in space polynomial in $\vert t \vert$, $\vert e \vert$ and $\vert \varphi \vert$~(ex. \cite{fionda2016complexity}). The model-checking of $\psi \in \ftllang$ against some $\langle t, \instance \rangle$ can be done by enumerating all relevant environments $e \in \objects^q$, and checking that $t, e \models \varphi$, where $\varphi$ is the quantifier-free part of $\psi$. As a consequence, the $\ftllang$ learning problem is in PSPACE. Even though this shows membership, the potential PSPACE-hardness of our problem is still an open problem.

\section{Planning problem preprocessing}
\label{sec:preprocessing}

% 
%\subsection{Pre-processing}

We present in this section the transformations we bring to the PDDL planning problem before it is passed to our algorithm for learning \ftllang{} formulas.
%As our algorithm is based on a compilation of the \ftllang{} learning problem into MaxSAT, reducing the size of the compiled form is crucial for it to run in reasonable time. 

%\arnaud{Add the Rovers example?}

% This section presents various ideas for modifying the planning model before we start encoding our learning problem into MaxSAT. The model thus obtained is a relaxation of the original instance, in the sense that there is a loss of information. However, we try to keep as much relevant bits of information as possible.

% \arnaud{This subsection is still a draft, only the main ideas are sketched}

\paragraph{Predicate splitting}

Each predicate is split into several predicates of size 2, in order to curb the number of fluents while conserving the links between pairs of objects. This allows us to synthesize formulas containing predicates of high arity, while keeping the number of quantifiers of the formula low.
%, for fluents that result of predicates of larger sizes, only a couple objects are important to understand the gist of the plan.

Concretely, a predicate of the form $p(x, y, z)$ will be split into newly-created predicates $p_{12}(x, y)$, $p_{13}(x, z)$, and $p_{23}(y, z)$. Predicate splitting leads to significantly fewer fluents than if the task was to be grounded as is: for a predicate of arity $n \geq 2$, to be grounded with instance $\instance$, there are $O(n^2 \vert \objects \vert^2)$ associated fluents, while there would be $O(\vert \objects \vert^n)$ if the predicate was not split.

Even though the planning model thus obtained is less rich than the original one, we argue that predicate splitting allows us to learn formulas that would be otherwise out of computational reach. %of our procedure.
%As will be shown in our experimental trials, learning formulas with 3 variables requires significant computational efforts. Thus, splitting predicates of arity 3 and more allows the information they bear to occur in the formula, even if it is incomplete.

%In the case where we try to split a predicate into predicates of a greater arity, nothing is done. For instance, trying to split a binary predicate into predicates of size 2 or 3 will leave the predicate as-is.

\paragraph{Goal predicates}
% The language $\ftllang$ naturally allows us to reason on the initial state. However, in its current state, it does not allow reasoning on the goal conditions, which depend on the instance and are trace-agnostic.

In order to allow the learnt formulas to reason on the goal state, we introduce \emph{goal predicates}. For every predicate $p \in \predicates$, we introduce the predicate $p^G$. Then, for each instance $\instance$, we introduce the \emph{latent state} $s_{\instance}$, which is intuitively a set of fluents that are true in every state of every trace associated to $\instance$. 

For every fluent $p(o_1, \ldots, o_{\arity{p}})$ of the goal state $\goal$ of $\instance$, we add the fluent $p^G(o_1, \ldots, o_{\arity{p}})$ to  $s_{\instance}$.

\section{Topology-based guiding}
\label{sec:topologyguiding}

\paragraph{TL chains}
An interesting representation for formulas $\varphi$ of $\tllang$ is a representation as \emph{TL chains}. They are the adaptation to our language of the notion of chain~\cite{knuth2020art,riener19exact}, which is useful for representing formulas of modal or propositional logic.

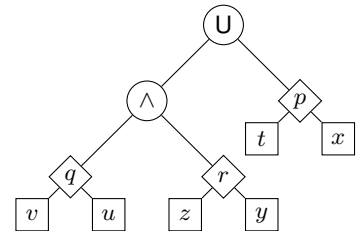
\begin{figure}
    \center
    \begin{tikzpicture}
        \node[circle, draw, inner sep=2pt] (n1) {$\textsf{U}$};
        \node[circle, draw, inner sep=2.5pt, below left of=n1, node distance=10ex] (n2) {$\wedge$};
        \node[shape aspect=1, diamond, draw, inner sep=1.5pt, below right of=n1, node distance=10ex] (p1) {$p$};
        \node[shape aspect=1, diamond, draw, inner sep=1.5pt, below left of=n2, node distance=10ex] (p2) {$q$};
        \node[shape aspect=1, diamond, draw, inner sep=2pt, below right of=n2, node distance=10ex] (p3) {$r$};
    
        \node[shape aspect=1, rectangle, draw, minimum size=3ex, below right of=p1, node distance=5ex] (v11) {$x$};
        \node[shape aspect=1, rectangle, draw, minimum size=3ex, below left of=p1, node distance=5ex] (v12) {$t$};
        \node[shape aspect=1, rectangle, draw, minimum size=3ex, below right of=p2, node distance=5ex] (v21) {$u$};
        \node[shape aspect=1, rectangle, draw, minimum size=3ex, below left of=p2, node distance=5ex] (v22) {$v$};
        \node[shape aspect=1, rectangle, draw, minimum size=3ex, below right of=p3, node distance=5ex] (v31) {$y$};
        \node[shape aspect=1, rectangle, draw, minimum size=3ex, below left of=p3, node distance=5ex] (v32) {$z$};
    
        \draw (p1) -- (n1) -- (n2) -- (p2);
        \draw (n2) -- (p3);
        \draw (v11) -- (p1) -- (v12);
        \draw (v21) -- (p2) -- (v22);
        \draw (v31) -- (p3) -- (v32);
        
    \end{tikzpicture}
    % \caption{A TL chain example, as well as a possible assignment of symbols to its nodes. The TL chain on the right has been assigned symbols to every one of its nodes, and represents the formula $(q(v,u) \wedge r(z, y)) \, \textsf{U} \, p(t, x)$}
    \caption{A TL chain example, which has been assigned symbols to its nodes. It represents the formula $(q(v,u) \wedge r(z, y)) \, \textsf{U} \, p(t, x)$}
    \vspace{3ex}
    \label{fig:tl_chain_example}
\end{figure}

A \emph{TL chain} is a Directed Acyclic Graph (DAG) which has three types of nodes: logical connector nodes (represented as $\circ$ in the example of Figure~\ref{fig:tl_chain_example}), predicate nodes (represented as $\diamond$) and variable nodes (represented as $\square$). In order to represent a correct \tllang{} formula, logical connector nodes can only be children of logical connector nodes, predicate nodes children of logical connector nodes, and variable nodes children of predicate nodes. We also impose that every leaf is a variable node. In addition, to stay consistent with the choices we made in Section~\ref{sec:preprocessing}, we only work with TL chains that are binary trees, whose inner nodes have exactly two children.

By assigning a symbol of the correct type (i.e., a logical connector, a predicate symbol or a variable) to each node, we end up with a representation of a \tllang{} formula, as illustrated in Figure~\ref{fig:tl_chain_example}. %Figure~\ref{fig:tl_chain_example} shows the representation as a TL chain of the formula $(q(v,u) \wedge r(z, y)) \, \textsf{U} \, p(t, x)$.

For each connector node $i$ of the TL chain, we will denote $\succl{i}$ (resp. $\succr{i}$) the left (resp. right) child of node $i$. It is guaranteed to exist, but might sometimes be a predicate node.
In the case of connectors $\alpha \in \{\neg, \bigcirc, \bbigcirc, \lozenge, \blozenge, \square, \bsquare\}$ that have arity 1, we will use the convention that the value of the right successor is ignored (and will not appear in the \ftllang{} formula that ensues), and the left successor will be the root of the formula under the operator $\alpha$.

% \begin{definition}
%     \label{def:ftlchain}
%     An \emph{FL-chain} is a DAG of depth at least 2, where:
%     \begin{itemize}
%         \item Leaves are called \emph{variable nodes}
%         \item Parents of leaves are called \emph{predicate nodes}
%         \item Other nodes are called \emph{logical connector nodes}
%         \item Inner nodes have exactly two children
%     \end{itemize}
% \end{definition}

In order to alleviate the pressure on the MaxSAT solver, we impose the topology of the output quantifier-free formula before encoding the problem into a propositional formula. This idea was first introduced in~\cite{riener19exact}, in the case of a search for LTL formulas. We also fix the quantifiers of the formula before the encoding, as well as the associated types. All that is left to the MaxSAT solver is to ``fill in the blanks'' in the TL chains that it is given, so that the associated \ftllang{} formula fits the input as well as possible. %In the next section, we present the encoding that allows it to do so.

% An interesting aspect of the constraints we impose on the form of the output formula, is that they force the algorithm to produce a wide diversity of formulas. % \arnaud{Prendre en compte les remarques à partir de cette phrase}

% \paragraph{FTL chain}
% We call \emph{FTL chain} any tuple $\langle Q, C\rangle$, where $Q$ is a sequence of quantifier symbols (i.e. symbols of $\{\forall, \exists\}$) and $C$ is a TL chain.

% In the rest of the article, for practical reasons, we restrict ourselves to formulas of the form $\forall x_1 \cdots \forall x_k \exists x_{k+1} \cdots \exists x_b \varphi$, where $\varphi$ is a quantifier-free FO-LTL$_f$ formula, for which every argument of every predicate is a variable $x_i$. 
% \arnaud{I think that we can still find a way to use constants too, through a correctly done preprocessing step. The only important thing is, no free variable.}

% \paragraph{Counting TL chains}
% As there is a one-to-one correspondence between TL chains with $r$ connector nodes and binary trees of size $r$, it can be shown that the number of TL chains with $r$ connector nodes is given by the $r$-th Catalan number $\mathcal{C}_r$, which is given by:

% \[
%   \catalan{r} = \frac{(2r)!}{(r+1)!\;r!}
% \]

% The Catalan series is known for its quick growth: the first few Catalan numbers, starting at $c = 0$, are 1, 1, 2, 5, 14, 42, 132, 429, 1430, 4862, 16796. As each TL chain is associated to a MaxSAT instance, this gives us an order of magnitude on the size of the formulas that are within the reach of our method.

\paragraph{Quantifiers}
In the rest of the article, for practical reasons, we restrict ourselves to learning formulas of the form $\forall x_1 \cdots \forall x_b \exists x_{b+1} \cdots \exists x_q \varphi$, where $\varphi$ is a formula of \tllang{}, for which every argument of every predicate is a variable $x_i$. This allows us to curb the size of the MaxSAT encoding. % This choice was made with the aim of limiting the number of MaxSAT instances to solve, but also to curb the size of the MaxSAT encoding.
%Indeed, as will be shown in Equation~(\ref{eq:env_selection_tsatvar}), 

\section{Reduction to MaxSAT}
\label{sec:encoding}

% In this section, we present the core of our contribution, which consists of a reduction of (a restricted version of) the \ftllang{} learning problem (Problem~\ref{prob:foltlf_learning}) to MaxSAT. Our encoding adapts parts of previous works found in the literature~\cite{gaglione2021learning,riener19exact,neider2018learning}.

\subsection{Learning algorithm}

Algorithm~\ref{alg:ftllearning} summarizes the procedure that we use to learn \ftllang{} formulas out of our input. The subroutines work as follows: gen\_TLchains($r$) enumerates every TL chain having exactly $r$ connectors. gen\_quantifiers($q$) enumerates sequences of quantifier symbols of size $q$, such that all universal quantifiers $\forall$ appear before existential quantifiers $\exists$. gen\_types($\domain$, $q$) enumerates every $q$-combination of types in the type tree $\types$ of $\domain$. Finally, the main subroutine, find\_formula($\domain, \traces, \rho, \{Q_i\}, \{\type{}_i\}, \tscore$), encodes the problem of finding an \ftllang{} formula fitting the instantiated traces of $\traces$, with the constraints imposed by the TL chain $\rho$, the quantifiers $\{Q_i\}$, and the types $\{\type{}_i\}$. find\_formula then returns (one of) the best formula(s) it finds, or the token FAIL if none is found.

\SetAlgoNoLine
\begin{algorithm}[h]
    \KwIn{Domain $\domain$, traces $\traces$, parameters $r, q$, and function $\tscore$}
    \KwOut{A set of \ftllang{} formulas}
    found\_formulas := [\,] \\
    \For{$\rho \in $ gen\_TLchains($r$)}{
        \For{$Q_1, \ldots, Q_q \in $ gen\_quantifiers($q$)}{
            \For{$\type_1, \ldots, \type_q \in $ gen\_types($\domain$, $q$)}{
                $\psi \leftarrow$ find\_formula($\domain, \traces, \rho, \{Q_i\}, \{\type{}_i\}, \tscore$)\;
                \lIf{$\psi \not=$ \text{FAIL}}{found\_formulas.add($\psi$)}
            }
        % Alt: uIf for a newline
        }
    }
    \Return{\text{found\_formulas}}
    \caption{\ftllang{} learning}
    \label{alg:ftllearning}
\end{algorithm}

\subsection{Preliminaries to the encoding}

\paragraph{Variables}
Our MaxSAT encoding is built on the set of variables that follows.
When possible, we use the following conventions, as closely as possible: nodes of the FL-chain are denoted by $i$ when they are logical connectors (represented by $\bigcirc$ in Figure~\ref{fig:tl_chain_example}), by $\ell$ when they are predicate nodes (represented by $\lozenge$), and by $v$ when they are first-order variable nodes (represented by $\square$). A trace is denoted by $t$, and a position in this trace is denoted by $k$ (i.e., the $k$-th state). Moreover, $j$ is an index for a variable of the quantifiers, and $p$ is a predicate.

This leads us to the following variables, as will be used in the MaxSAT encoding. Greek letters denote decision variables while latin characters are for ``technical'' variables.
\begin{itemize}[label=$\circ$]
    \item \tvalenvvar{i}{t}{k}{e}: In position $k$ of trace $t$, with environment $e$, the formula rooted at node $i$ is true.
    \item \qtfvar{j}{v}{\ell}: The $v$-th variable of predicate node $\ell$ is the variable of quantifier $j$.
    \item \predfvar{\ell}{p}: The predicate of node $\ell$ is $p$.
    % \item \qtcvar{j}{c}: The object class on which quantifier $j$ ranges is $c$.
    \item \logcvar{i}{q}: The logical connector at node $i$ is $q$.
    \item \tsatvar{t}: Trace $t$ is currently satisfied by the first order formula 
\end{itemize}
%
% \arnaud{Remark:} Instead of $\qtfvar{j}{v}{\ell}$, we could directly consider variables that select the whole set of quantifiers. Variables $\qtfvar{j}{v}{\ell}$ only occur in equation~\ref{eq:env_selection}, and with Tseitin's transformation, they would actually create the variable evoked in the previous sentence. So we might as well create them directly.

\paragraph{``Exactly one'' constraints}
In the encoding of a problem into SAT, some situations require that \emph{exactly one} variable, out of a set of variables, is true. Efficient encodings have been widely studied: see for instance~\cite{holldobler2013efficient,nguyen2015new}. In the following, we will denote \exactlyone{s \in S}($v_s$) the set of propositional constraints enforcing that exactly one of the variables of $\{v_s \mid s \in S\}$ is true.

%In particular, the authors propose the \emph{bimander} encoding, that seems to often be easier to solve for SAT solvers than the naive encoding.

% In the encoding of a problem into SAT, some situations require that \emph{at most one} variable, out of a set of variables, is true. There exist encodings that are more efficient than the naive one to define such ``at most one'' constraints: see for instance~\cite{holldobler2013efficient,nguyen2015new} for a survey on these encodings. Like many other solvers, the MaxSAT solver that we use proposes a built-in function for this. More specifically, it offers a built-in function for \emph{exactly one} constraints, where exactly one variable of a set must be true in a model of the formula.
% %In particular, the authors propose the \emph{bimander} encoding, that seems to often be easier to solve for SAT solvers than the naive encoding.

% In the following, we will denote \exactlyone{s \in S}($v_s$) the set of propositional constraints enforcing that at most one of the variables of $\{v_s \mid s \in S\}$ is true.

% \arnaud{Maybe that something like \atmostone{}($\{v_s \mid s \in S\}$) is more formal? But is it more readable?}

\subsection{Core constraints}

Some of the constraints below are adapted from~\cite{gaglione2021learning,riener19exact,neider2018learning}, which are concerned with LTL. Our main contribution is the adaptation of the encoding to our language \ftllang{}, which differs from LTL by its tighter links with PDDL planning models through first-order components.

In the following, we suppose that an empty TL chain $\rho$ has been computed, and that the associated quantifiers and types have been decided. We will denote $n$ its number of connector nodes, and $m$ its number of predicate nodes. As a consequence, there are $2m$ variable nodes. As previously, the number of quantifiers is denoted $q$. The first $b \leq q$ quantifiers are universal, while the others are existential.

We also suppose that the types on which the quantifiers range, denoted $\type_1, \ldots, \type_q$, are already chosen. As a consequence, in this section, the set of relevant environments for instance $\instance_t$ associated to trace $t$, denoted $\env{}_{\instance_t}$, only consists of environments of the form $\{x_u := o_u\}_{1 \leq u \leq q}$ where, $\type{}(o_u) = \type_u$, for $u \in \llbracket 1, q\rrbracket$.

\paragraph{Syntactic constraints}

This section describes the constraints that ensure that the formula is syntactically well-formed.
%In our framework, we consider that we already have an empty FL-chain $C$ (that is, an FL-chain where nodes are empty), and we denote $b$ the number of quantifiers, $n$ the number of \emph{logical connector} nodes, $m$ the number of predicate nodes. As a consequence, there are $2m$ arguments.

% \arnaud{Also specify that the type of each quantifier is chosen a priori (ie. we will have 2 $\forall$ then 2 $\exists$, for instance)}

The following constraints respectively ensure that every logical connector node has exactly one logical connector assigned, that every predicate node has exactly one predicate, and that each argument of each predicate is bound to a variable on which the formula quantifies. %Recall that $\conset$ is the set of all logical operators.%, be them propositional or temporal.
\begin{equation*}
    \bigwedge_{i \leq n} \exactlyone{c \in \Lambda}(\logcvar{i}{c}) \wedge \bigwedge_{\ell \leq m} \exactlyone{p \in \predicates}(\predfvar{\ell}{p})
\end{equation*}
\begin{equation*}
    \bigwedge_{\ell \leq m} \bigwedge_{s \in \{1, 2\}} \exactlyone{j \leq b}(\qtfvar{j}{s}{\ell})
\end{equation*}
% \arnaud{Changed this equation to include latent predicates}
% \begin{equation}
%     \atmostone{}(\{\predfvar{\ell}{p} \mid p \in \predicates{} \setminus \latpredicates{}\} \cup \{\lpredfvar{\ell}{p} \mid p \in \latpredicates{}\})
% \end{equation}

% Similarly, even though quantifiers are chosen with the topology of the formula (i.e. along with the FL-chain), the objects classes on which they quantify are to be determined by the solver. The existence of a single class is ensured through the following, where we recall that $\objclassesset{}$ is the set of all possible objects classes:

% \begin{equation}
%     \bigwedge_{j \leq b} \bigvee_{c \in \objclassesset{}} \qtcvar{j}{c}
% \end{equation}

% \begin{equation}
%     \bigwedge_{j \leq b} \atmostone{c \in \objclassesset{}}(\qtcvar{j}{c})
% \end{equation}

% \begin{equation}
%     \bigwedge_{\ell \leq m} \bigwedge_{s \in \{1, 2\}} \bigvee_{j \leq b} \qtfvar{j}{s}{\ell}
% \end{equation}

\paragraph{Semantic constraints}
These constraints ensure that the formula found by the solver is consistent with the traces, and is reminiscent of the model-checking algorithm for modal logic.

The following clauses ensure that the formula $\psi$ that is synthesized is consistent with the traces of $\traces$. This is made in accordance with the environments imposed by the quantifier, which are iterated upon. The variable $\tsatvar{t}$ is true iff for every required environment $e$, $\varphi\left[e\right]$ is satisfied by $t$ (where $\varphi\left[e\right]$ is the evaluation of formula $\varphi$ in environment $e$, and $\varphi$ is the quantifer-free part of the formula we synthesize). Thus, for every trace $t \in \traces$, we add the following:
%
% We suppose that the first $k$ quantifiers are universal $\forall$ quantifiers, while the next ones are existential $\exists$. The idea is to start by deciding which objects classes to quantify on, and considering all environments made of combinations of objects that comply with the objects classes.
%
% Now suppose that we have chosen a series of objects classes $\objclasssymb_1, \ldots, \objclasssymb_q$. For each such combination, we add the following clauses (for which the first conjunction ranges over \emph{positive} traces):
% \arnaud{Change the notations for the objects classes, we have objects here that have the same $o$ letter assigned to them}
%
% \begin{equation}
%     \label{eq:env_selection}
%     \bigwedge_{t \in \postraces} (\bigwedge_{j \leq b} \qtcvar{j}{\objclasssymb{}_j}) \Rightarrow \bigwedge_{o_1 \in \objclasssymb{}_1} \cdots \bigwedge_{o_k \in \objclasssymb{}_k} \bigvee_{\substack{o_{k+1} \\ \in \objclasssymb{}_{k+1}}} \cdots \bigvee_{o_q \in \objclasssymb{}_q} \tvalenvvar{1}{t}{1}{\{x_1 := o_1, \ldots, x_q := o_q\}}
% \end{equation}
%
% which can be rewritten in the following form, so that clauses become more apparent:
%
% \begin{equation*}
%     \bigwedge_{t \in \postraces} \bigwedge_{o_1 \in \objclasssymb{}_1} \cdots \bigwedge_{o_k \in \objclasssymb{}_k} (\bigwedge_{j \leq b} \qtcvar{j}{\objclasssymb{}_j})  \Rightarrow \bigvee_{\substack{o_{k+1} \\ \in \objclasssymb{}_{k+1}}} \cdots \bigvee_{o_q \in \objclasssymb{}_q} \tvalenvvar{1}{t}{1}{\{x_1 := o_1, \ldots, x_q := o_q\}}
% \end{equation*}
%
% \arnaud{Formula where the $\tsatvar{t}$ variables are used:}
%
\begin{equation}
    \label{eq:env_selection_tsatvar}
    \tsatvar{t} \Leftrightarrow
    %\left(
    \bigwedge_{\substack{o_1 \in \objclasssymb{}_1 \\ \cdots \\ o_k \in \objclasssymb{}_k}} \bigvee_{\substack{o_{k+1} \in \objclasssymb{}_{k+1} \\ \cdots \\ o_q \in \objclasssymb{}_q}} \tvalenvvar{1}{t}{1}{\{x_u := o_u\}_{1 \leq u \leq q}}
    %\right)
\end{equation}
%
%In the case of a negative trace, the clause is almost exactly the same, only that the variable $\tvalvar{1}{t}{1}$ is preceded by a negation $\neg$.
%In the case of a negative trace, the encoding is the exact same: the only point that will differ is that the weight of the clause will have the opposite sign, as we detail in the corresponding paragraph further down. In the case where on wishes to encode the learning problem as a SAT problem instead of a MaxSAT problem, a negation $\neg$ should be added in front of the decision variable $\qtcvar{j}{\objclasssymb{}_j}$.
% In addition, note that one could also consider any alternation of quantifiers, but the encoding would be far bigger, as the formula needs to be translated into a CNF.
%
% The next equations ensure that the truth values of the formulas rooted at each node of the FL-chain are correctly interpreted.
%
The following constraints ensure that formulas that consist of a single literal (i.e., a positive or negative fluent) are consistent with the \tvalvarsymb{} variables, that give the truth value of a trace at a certain position in the trace, at each node of the TL chain.

% \arnaud{Introduce the $\traces$ notation for $\postraces \cup \negtraces$.} \arnaud{Also we should add dummy predicates $\neg p$ to allow negations of fluents}.

Such constraints appear once for every trace $t \in \traces$, for every position $k \leq \vert t \vert$ of this trace, for every predicate node $\ell \leq m$ and every predicate $p \in \predicates$, for every pair of quantifiers (positions) $j_1, j_2 \leq q$, and for each relevant environment $e \in \env{}_{\instance_t}$.
\begin{equation}
    \label{eq:predicate_truth}
    \predfvar{\ell}{p} \wedge \qtfvar{j_1}{1}{\ell} \wedge \qtfvar{j_2}{2}{\ell} \Rightarrow 
      \begin{cases}
        \tvalenvvar{\ell}{t}{k}{e} & \text{if } t[k] \models p(x_{j_1}, x_{j_2})[e]\\
        \neg \tvalenvvar{\ell}{t}{k}{e} & \text{otherwise}
      \end{cases} 
\end{equation}
%
% However, the above formula is slightly different in the case where $p$ ranges over the marked predicates:
%
% \begin{equation*}
%     \bigwedge_{t \in \traces} \bigwedge_{k \leq \vert t \vert} \bigwedge_{\ell \leq m} \bigwedge_{p^G \in \predicates} \bigwedge_{j_1 \leq q} \bigwedge_{\substack{j_2 \leq q \\ j_1 \not= j_2}} \bigwedge_{e \in E} \predfvar{\ell}{p} \wedge \qtfvar{j_1}{1}{\ell} \wedge \qtfvar{j_2}{2}{\ell} \Rightarrow 
%       \begin{cases}
%         \tvalenvvar{\ell}{t}{k}{e} & \text{if } G \models p(x_{j_1}, x_{j_2})[e]\\
%         \neg \tvalenvvar{\ell}{t}{k}{e} & \text{otherwise}
%       \end{cases} 
% \end{equation*}
% \arnaud{Change the way it is defined. There's a $\cdot^G$ in the $\bigwedge$, it's not so nice. }
%
% \arnaud{Note that these formulas can be simplified as the truth value of $\tvalenvvar{\ell}{t}{k}{e}$ is known} => Well, not at all...
%
Constraints (\ref{eq:neg_semantics}) to (\ref{eq:finally_semantics}) appear once for each connector node $i \leq n$ of the formula, each position $k \leq \vert t \vert$ of each trace $t \in \traces$, and for each environment $e \in \env{}_{\instance_t}$. They ensure that the logical operators are correctly interpreted.

In the case where the logical connector at node $i$ is a negation $\neg$, or $\Delta \in \{\wedge, \vee, \Rightarrow\}$, or the next operator $\bigcirc$, we have: 
%\arnaud{Maybe I'll remove this to only allow negation normal form? This would make redundant a lot of clauses, so it's good}
%
\begin{equation}
    \label{eq:neg_semantics}
    \logcvar{i}{\neg} \Rightarrow \left(\tvalenvvar{i}{t}{k}{e} \Leftrightarrow \neg \tvalenvvar{\succl{i}}{t}{k}{e} \right)
\end{equation}
\begin{equation}
    \logcvar{i}{\Delta} \Rightarrow \left(\tvalenvvar{i}{t}{k}{e} \Leftrightarrow \left( \tvalenvvar{\succl{i}}{t}{k}{e} \; \Delta \; \tvalenvvar{\succr{i}}{t}{k}{e} \right) \right)
\end{equation}
% 
% In the case of the next operator $\bigcirc$, we have the following:
% \arnaud{I had something more to say about this}
% \arnaud{Note that we have to create dummy variables $\tvalenvvar{i}{t}{\mid t\mid + 1}{e}$ that are always false. They can be directly replaced in the encoding.}
%
\begin{equation}
    \logcvar{i}{\bigcirc} \Rightarrow \left(\tvalenvvar{i}{t}{k}{e} \Leftrightarrow \tvalenvvar{\succl{i}}{t}{k+1}{e} \right)
\end{equation}

with the convention that $\tvalenvvar{\succl{i}}{t}{\mid t \mid+1}{e}$ is replaced by $\bot$ during the encoding itself.
In the case of the finally operator $\lozenge$:% and $\blozenge$:
\begin{equation}
    \label{eq:finally_semantics}
    \logcvar{i}{\lozenge} \Rightarrow \left(\tvalenvvar{i}{t}{k}{e} \Leftrightarrow \bigvee_{\substack{k' \\ k \leq k' \leq \mid t \mid}} \tvalenvvar{\succl{i}}{t}{k'}{e} \right)
\end{equation}
%
% \begin{equation}
%     \logcvar{i}{\blozenge} \Rightarrow \left(\tvalenvvar{i}{t}{k}{e} \Leftrightarrow \bigvee_{\substack{k' \\ 1 \leq k' \leq k}} \tvalenvvar{\succl{i}}{t}{k'}{e} \right)
% \end{equation}

% \arnaud{Can be skipped if space is needed}
% In the case of $\square$:% and $\bsquare$:
% % \begin{subequations}
% \begin{equation}
%     \label{eq:globally_semantics}
%     \logcvar{i}{\square} \Rightarrow \left(\tvalenvvar{i}{t}{k}{e} \Leftrightarrow \bigwedge_{\substack{k' \\ k \leq k' \leq \mid t \mid}} \tvalenvvar{\succl{i}}{t}{k'}{e} \right)
% \end{equation}
%
% \begin{equation}
%     \logcvar{i}{\bsquare} \Rightarrow \left(\tvalenvvar{i}{t}{k}{e} \Leftrightarrow \bigwedge_{\substack{k' \\ 1 \leq k' \leq k}} \tvalenvvar{\succl{i}}{t}{k'}{e} \right)
% \end{equation}
% \end{subequations}

The case of the temporal operators $\square$, $\bbigcirc$, $\blozenge$, $\bsquare$ and $\textsf{U}$ can be encoded in a way that is similar to the constraints above.

% In the case of $\textsf{U}$:

% \begin{equation}
%     \logcvar{i}{\textsf{U}} \Rightarrow \left(\tvalenvvar{i}{t}{k}{e} 
%     \Leftrightarrow 
%         \bigvee_{\substack{k' \\ k \leq k' \leq \mid t \mid}} \left( 
%             \tvalenvvar{\succr{i}}{t}{k'}{e}
%             \wedge 
%             \bigwedge_{\substack{k'' \\ k \leq k'' < k'}} 
%                 \tvalenvvar{\succl{i}}{t}{k''}{e} 
%         \right) 
%     \right)
% \end{equation}

% Note that since this is a strong Until, for any formula $p \textsf{U} q$, we require $q$ to be eventually true. Hence $k'$ ranging until $\mid t \mid$ \emph{not included}.

\paragraph{Well-formed fluents constraints}
The following constraints ensure that, in the output formula $\psi$, there is a consistency between the types of the variables and the arguments of predicates are assigned to. In other words, when a variable $x$ of type $\type$ is chosen to be the $v$-th argument of a predicate $p$ that occurs in $\psi$, we require that $\type = \predvartype{p}{v}$. This can be done through the following constraints:

\begin{equation}
    \bigwedge_{j \leq q} \bigwedge_{\ell \leq m} \bigwedge_{p \in \predicates} \bigwedge_{j \leq q} \bigwedge_{\substack{v \leq 2 \\ \predvartype{p}{v} \not= \type_j}} \neg \predfvar{\ell}{p} \vee \neg \qtfvar{j}{v}{\ell}
\end{equation}

\paragraph{Weights for the MaxSAT solver}

Recall that we wish to find a formula $\psi$ that maximizes the function given in Problem~\ref{prob:foltlf_learning}.
%
% \[
%     \sum_{\langle t, \instance{} \rangle \in \traces} \hspace{1ex} \tscore(\langle t, \instance \rangle)\left[ \langle t, \instance \rangle \models \psi \right]
% \]
%
The objective of the MaxSAT solver is to minimize the total weight of the falsified soft clauses. As such, for each instantiated trace $\langle t, \instance \rangle$, we add the clause $s_t$, with weight $\tscore{}(\langle t, \instance \rangle)$. This penalizes formulas that falsify traces with a positive score, while rewarding formulas that falsify traces with a negative score.

\paragraph{Pruning non-discriminatory formulas}
With a given configuration of TL chain, quantifiers and types, it is not guaranteed that there exists a formula $\psi$ that captures (some of) the positive traces while falsifying (some of) the negative traces. To prevent tautology or unsatisfiable formulas from occurring, we enforce the constraint that at least one positive trace is captured and one negative trace is not.
% Without further precautions, our algorithm can output formulas that are true on all traces, or false on all traces. These formulas are often tautologies or unsatisfiable, and still have a non-zero score as they completely capture the positive or negative traces.

%Sometimes, for a given FTL Chain, no formula fitting that chain can be true on some positive trace while being false on some negative trace. Since it's not interesting \arnaud{Explain why}, we add the following clauses, to ensure that the formula found is not too lax by being a tautology or unsatisfiable.

% The following clauses ensure that at least one positive trace and one negative trace are captured:
%
% \begin{equation}
% \bigvee_{\substack{t \in \traces \\ \tscore(t) \geq 0}} \tsatvar{t} \wedge \bigvee_{\substack{t \in \traces \\ \tscore(t) < 0}} \neg \tsatvar{t}
% \end{equation}
%
\subsection{Formula quality enhancement}

The constraints of this section filter the solutions so that less interesting formulas, or formulas that could be computed by a run of our algorithm with smaller parameters, are barred from being output.

\paragraph{Syntactic redundancies prevention} 
These constraints prevent idempotent and involutive modalities and operators from being chained in the output formula. These include the negation $\neg$, as well as the temporal operators $\lozenge$ (for which $\lozenge \lozenge \varphi \equiv \lozenge \varphi$) and $\square$ (which is, likewise, idempotent).
%In order to prevent the operator $\alpha \in \{\neg, \lozenge, \blozenge, \square, \bsquare\}$ from appearing in a node of the TL chain and its left-successor, we add the following constraints, when possible (i.e. when both $i$ and $\succl{i}$ are defined):
%
% \begin{equation}
%     \neg \logcvar{i}{\alpha} \vee \neg \logcvar{\succl{i}}{\alpha}
% \end{equation}
%
%The constraints above improve the quality of the formulas found, while being relatively easy to satisfy, as they make the most of unit propagation.

In addition, we prevent redundancies of the form $p(x, y) \, \Delta \, p(x, y)$, where $\Delta \in \{ \wedge, \vee, \textsf{U}, \Rightarrow \}$ is a binary operator. In every case, there exists a smaller (sub-)formula that can be found and that expresses the same property, without the redundant atom.
For space reasons, we skip the presentation of the constraints.

\paragraph{Variable visibility}

%As the size of the encoding is exponential in the number $q$ of expected quantifiers,
We wish to ensure that every variable that we quantify upon in the output formula $\psi$ also appears in an atom of $\psi$. Otherwise, an equivalent formula could be found by running the algorithm with fewer quantifiers. This is why we force each variable to appear at least once in some atom.

\section{Experiments}
\label{sec:experiments}

We implemented Algorithm~\ref{alg:ftllearning} in Python 3.10, using the MaxSAT solver Z3~\cite{deMoura2008z3}. Experiments were conducted on a machine running Rocky Linux 8.5, powered by an Intel Xeon E5-2667 v3 processor, with a 9-hours cutoff and using at most 8GB of memory per run. The code of our implementation and our data are available online~\cite{lequen2024implementation}. % It also includes tools to evaluate formulas against a dataset and a planning model.
Additional, more detailed test results can also be found in~\cite{lequen2024implementation}.

Even though we often managed to quickly find a formula that perfectly captures the set of examples, we let the algorithm run to the end, so that all TL chains and combinations of quantifiers and types are enumerated.% at least once.

\paragraph{Building the data sets}
% \arnaud{Say that we omitted certain object classes, for instance the ``objects'' OC}
To assess the performances of our algorithm, we considered domains from the International Planning Competition (IPC), 2 of which are described in Section~\ref{subsec:learnt_formulas}. For each of these domains, we generated 23 instances that model problems with similar goals. Then, for each domain, we designed three domain-specific planners, that solve the tasks in a distinctive way.

% We then used 5 planners from the IPC to generate plans for each instance, that our algorithm converted to traces. On average, plans had 10.2 operators, but some instances include plans of size up to 23.

We built our training sets by selecting 3 small planning instances of each domain, and the associated traces for each planner \--- for a total of 9 traces per domain. We then created the tasks of finding a formula recognizing the behaviour of each planner, out of 1, 2 or 3 of the training instances. The 20 remaining instances (and their associated plans) were used in the test set. The traces of our training set have length 5 to 21, with an average of 11.8 states.

\subsection{Examples of learnt formulas}
\label{subsec:learnt_formulas}

\begin{table}[t]
    \caption{Proportion of traces correctly classified by the most accurate formula on our test set, expressed in percentage (\%). Each table presents a different agent solving a set of Childsnack problems. \emph{\# ins.} indicates the number of instances in the training set, $q$ the number of quantifiers allowed, and $\vert \varphi \vert$ the number of logical operators allowed. Dots correspond to configuration where the run did not terminate.}
    \center
    \vspace{-1ex}
    \begin{tabular}{r||rr|rr|rr}
        \multicolumn{7}{c}{\textbf{GS}} \\
        \# ins. & \multicolumn{2}{c}{1} & \multicolumn{2}{c}{2} & \multicolumn{2}{c}{3} \\
        $q$ & \multicolumn{1}{c}{1} & \multicolumn{1}{c}{2} & \multicolumn{1}{c}{1} & \multicolumn{1}{c}{2} & \multicolumn{1}{c}{1} & \multicolumn{1}{c}{2} \\
        \hline
        $\vert \varphi \vert =$ 2 & 67 & 100 & 67 & 100 & 67 & 100 \\
        3 & 100 & 100 & 100 & 100 & 100 & 100 \\
        4 & . & 100 & . & 100 & . & .
    \end{tabular}

    \vspace{1ex}

    \begin{tabular}{r||rr|rr|rr}
        \multicolumn{7}{c}{\textbf{NGF}} \\
        \# ins. & \multicolumn{2}{c}{1} & \multicolumn{2}{c}{2} & \multicolumn{2}{c}{3} \\
        $q$ & \multicolumn{1}{c}{1} & \multicolumn{1}{c}{2} & \multicolumn{1}{c}{1} & \multicolumn{1}{c}{2} & \multicolumn{1}{c}{1} & \multicolumn{1}{c}{2} \\
        \hline
        $\vert \varphi \vert =$ 2 & 67 & 67 & 67 & 67 & 67 & 67 \\
        3 & 100 & 100 & 100 & 100 & 100 & 67 \\
        4 & . & . & . & . & . & .
    \end{tabular}

    \vspace{1ex}

    \begin{tabular}{r||rr|rr|rr}
        \multicolumn{7}{c}{\textbf{NGL}} \\
        \# ins. & \multicolumn{2}{c}{1} & \multicolumn{2}{c}{2} & \multicolumn{2}{c}{3} \\
        $q$ & \multicolumn{1}{c}{1} & \multicolumn{1}{c}{2} & \multicolumn{1}{c}{1} & \multicolumn{1}{c}{2} & \multicolumn{1}{c}{1} & \multicolumn{1}{c}{2} \\
        \hline
        $\vert \varphi \vert =$ 2 & 100 & 100 & 100 & 100 & 100 & 100 \\
        3 & 100 & 100 & 100 & 100 & 100 & 100 \\
        4 & . & . & . & . & . & .
    \end{tabular} 

    \label{tab:best_scores_childsnack}
\end{table}

\begin{table}[t]
    \caption{Total number of formulas found for Childsnack.}
    \center
    \vspace{-1ex}
    \begin{tabular}{r||rr|rr|rr}
        \multicolumn{7}{c}{\textbf{GS}} \\
        \# ins. & \multicolumn{2}{c}{1} & \multicolumn{2}{c}{2} & \multicolumn{2}{c}{3} \\
        $q$ & \multicolumn{1}{c}{1} & \multicolumn{1}{c}{2} & \multicolumn{1}{c}{1} & \multicolumn{1}{c}{2} & \multicolumn{1}{c}{1} & \multicolumn{1}{c}{2} \\
        \hline
        $\vert \varphi \vert =$ 2 & 22 & 110 & 22 & 112 & 22 & 123 \\
        3 & 92 & 624 & 92 & 624 & 92 & 671 \\
        4 & . & 746 & . & 757 & . & .
    \end{tabular} 

    \vspace{1ex}

    \begin{tabular}{r||rr|rr|rr}
        \multicolumn{7}{c}{\textbf{NGF}} \\
        \# ins. & \multicolumn{2}{c}{1} & \multicolumn{2}{c}{2} & \multicolumn{2}{c}{3} \\
        $q$ & \multicolumn{1}{c}{1} & \multicolumn{1}{c}{2} & \multicolumn{1}{c}{1} & \multicolumn{1}{c}{2} & \multicolumn{1}{c}{1} & \multicolumn{1}{c}{2} \\
        \hline
        $\vert \varphi \vert =$ 2 & 23 & 126 & 23 & 126 & 23 & 140 \\
        3 & 92 & 670 & 92 & 679 & 94 & 723 \\
        4 & . & 797 & . & 858 & . & .
    \end{tabular} 

    \vspace{1ex}

    \begin{tabular}{r||rr|rr|rr}
        \multicolumn{7}{c}{\textbf{NGL}} \\
        \# ins. & \multicolumn{2}{c}{1} & \multicolumn{2}{c}{2} & \multicolumn{2}{c}{3} \\
        $q$ & \multicolumn{1}{c}{1} & \multicolumn{1}{c}{2} & \multicolumn{1}{c}{1} & \multicolumn{1}{c}{2} & \multicolumn{1}{c}{1} & \multicolumn{1}{c}{2} \\
        \hline
        $\vert \varphi \vert =$ 2 & 7 & 76 & 7 & 78 & 8 & 92 \\
        3 & 79 & 604 & 79 & 681 & 82 & 725 \\
        4 & . & 784 & . & 845 & . & .
    \end{tabular} 

    \label{tab:formula_count_childsnack}
\end{table}

\begin{table}
    \caption{Total computation times for Childsnack (hh:mm:ss). Rows correspond to the number of allowed logical connectors. Dots correspond to configuration where the run did not terminate.}
    \center
    \vspace{-1ex}
    \begin{tabular}{r||rr|rr|rr}
        \multicolumn{7}{c}{\textbf{GS}} \\
        \# ins. & \multicolumn{2}{c}{1} & \multicolumn{2}{c}{2} & 
\multicolumn{2}{c}{3} \\
        $q$ & \multicolumn{1}{c}{1} & \multicolumn{1}{c}{2} & 
\multicolumn{1}{c}{1} & \multicolumn{1}{c}{2} & \multicolumn{1}{c}{1} & 
\multicolumn{1}{c}{2} \\
        \hline
        2 & 0:00:08 & 0:05:03 & 0:00:08 & 0:16:43 & 
0:01:14 & 0:44:48 \\
        3 & 0:00:35 & 0:19:53 & 0:00:35 & 1:07:27 & 0:06:46 & 3:35:18 \\
        4 & . & 0:25:27 & . & 1:32:07 & . & .
    \end{tabular}

    \vspace{1ex}

    \begin{tabular}{r||rr|rr|rr}
        \multicolumn{7}{c}{\textbf{NGF}} \\
        \# ins. & \multicolumn{2}{c}{1} & \multicolumn{2}{c}{2} & \multicolumn{2}{c}{3} \\
        $q$ & \multicolumn{1}{c}{1} & \multicolumn{1}{c}{2} & \multicolumn{1}{c}{1} & \multicolumn{1}{c}{2} & \multicolumn{1}{c}{1} & \multicolumn{1}{c}{2} \\
        \hline
        2 & 0:00:08 & 0:05:08 & 0:00:08 & 0:17:07 & 
0:01:09 & 0:47:51 \\
        3 & 0:00:37 & 0:20:13 & 0:00:37 & 1:11:13 & 0:07:05 & 3:44:11 \\
        4 & . & 0:26:29 & . & 1:33:27 & . & .
    \end{tabular}

    \vspace{1ex}

    \begin{tabular}{r||rr|rr|rr}
        \multicolumn{7}{c}{\textbf{NGL}} \\
        \# ins. & \multicolumn{2}{c}{1} & \multicolumn{2}{c}{2} & \multicolumn{2}{c}{3} \\
        $q$ & \multicolumn{1}{c}{1} & \multicolumn{1}{c}{2} & \multicolumn{1}{c}{1} & \multicolumn{1}{c}{2} & \multicolumn{1}{c}{1} & \multicolumn{1}{c}{2} \\
        \hline
        2 & 0:00:06 & 0:04:55 & 0:00:06 & 0:14:59 & 
0:00:48 & 0:40:28 \\
        3 & 0:00:31 & 0:19:00 & 0:00:31 & 1:07:48 & 0:05:42 & 3:14:13 \\
        4 & . & 0:25:20 & . & 1:30:16 & . & .
    \end{tabular} 

    \label{tab:total_computation_time_childsnack}
\end{table}

% \arnaud{REVENIR ICI LORSQUE: 1. LES TESTS SONT TERMINÉS 2. J'AI DÉCIDÉ COMMENT INTRODUIRE LE PROBLÈME, ET QUEL DOMAIN INTRODUIRE EN EXEMPLE FILÉ}

% In this section, we present some formulas that have been learnt by our algorithm over two domains.
%For each of these, we designed three domain-specific planners that solve the task in a distinctive way. We then built sets of plans in a similar way as in the previous section.

% For the first two domains, since the plans found by the 5 planners were all very similar, we handcrafted various agents that tackle the problem in a distinctive way. We kept off-the-shelf planners for the last domain. We then built sets of plans in a similar way as in the previous section.

\paragraph{Childsnack}
We designed three different agents that solve Childsnack instances, as introduced in Section~\ref{sec:ftllang}. Agents NGF and NGL compute solution plans of minimal size, and differ in that agent NGF makes sandwiches with \emph{no gluten first}, and agent NGL makes sandwiches with \emph{no gluten last}. Both agents make all sandwiches, put them on a tray, then serve the children. Agent GS greedily serves children: as soon as a sandwich is made, it is put on a tray and brought to a child. It also prioritizes gluten-free sandwiches.

For each of these behaviours, the total computation time and the total number of formulas found are depicted in Table~\ref{tab:total_computation_time_childsnack} and Table~\ref{tab:formula_count_childsnack}, respectively. As shown on Table~\ref{tab:best_scores_childsnack}, we manage to learn formulas that perfectly capture the test set, even when the number of examples is very limited, and when few quantifiers and logical operators are allowed. Formula~\ref{eq:notprepareduntillastminute}, given in Section~\ref{sec:ftllang}, is an example of a concise formula that describes the policy of agent GS perfectly. Other formulas include the following:
% In every instance, 2 trays are initially in the kitchen. The only difference between instances is in the number of children to serve, the smallest having 2. This is, however, enough to learn a wide variety of formulas that perfectly recognize the behavior of agent GS (among others) on our test set. Such formulas include, for instance, the following:
%
\begin{align}
    \forall x \in \text{Kitchen}. \, \exists y \in \text{Tray}. \, \lozenge (\text{at}(y, x) \wedge \blozenge \neg \text{at}(y, x)) \label{lf:childsnacks1} %\\
    %\forall x \in \text{Kitchen}. \, \exists y \in \text{Tray}. \, \lozenge (\neg \text{at}(y, x) \wedge \bigcirc \text{at}(y, x)) \label{lf:childsnacks2}
\end{align}
Formula (\ref{lf:childsnacks1}) expresses that agent GS eventually comes back to the (only) kitchen with some tray $y$, even though the tray was brought out of the kitchen at some point in the past. %Formula (\ref{lf:childsnacks2}) expresses the same idea, but pinpoints the moment when a tray is brought back to the kitchen.
The formula perfectly captures our test set, but does not perfectly capture our \emph{training} set. Indeed, the smallest instance of our training set contains as many children as there are trays, and thus, no tray has to be brought back to the kitchen. Our use of a reduction to MaxSAT allows us to be resilient to this kind of edge cases, and the formula above is satisfactory despite not perfectly fitting the training set.

% Even though our algorithm managed to learn concise formulas that perfectly capture agent NGL's behaviour, it failed to find in reasonable time a formula that discriminates agent NGF's behaviour with reasonable accuracy.

% ∃yϵSANDWICH.◯ ((◯ (no_gluten_sandwich(y))) ∧ (notexist(y)))

%\arnaud{Examples of formulas that recognize the other agents?}

% An agent keeps going back and forth from the kitchen to the tables, while the other puts everything on the tray first and then only goes to the tables.

\paragraph{Spanner}
Instances of the Spanner domain involve an operator that has to go from a shed to a gate to tighten some nuts, passing through a sequence of locations where single-use spanners can be picked up. Once a location is left, it can not be returned to. Thus, collecting enough spanners before reaching the gate is essential. % for solving the problem.

We developed three different behaviours for this domain. Agent ALL picks every possible spanner on its way to the gate, while agent SME picks exactly as many spanners as are needed to tighten the nuts at the gate. Agent SGL takes a single spanner and rushes to the gate, and can then only tighten one nut. Our algorithm learnt the following formulas, that perfectly recognize agent ALL:
%
%Among the formulas that we learnt, and that perfectly recognize plans belonging to agent ALL, we have the following:
%
\begin{align}
    &\forall x \in \text{Spanner}.\, \exists y \in \text{Operator}.\, \lozenge \, \square \, \text{carrying}(y, x) \label{lf:spanner1} \\
% (0) ∀xϵSPANNER.∃yϵMAN.◊ (□ (carrying(y,x)))
    &\forall x \in \text{Spanner}.\, \exists y \in \text{Location}.\, \text{at}(x, y) \wedge \lozenge \neg \text{at}(x, y) \label{lf:spanner2}
% ∀xϵSPANNER.∃yϵLOCATION.(at(x,y)) ∧ (◊ (¬(at(x,y))))  
\end{align}
Formula (\ref{lf:spanner1}) expresses that every spanner will be picked up by the (only) operator and carried for the rest of the plan, and Formula ($\ref{lf:spanner2}$) expresses that every spanner will be moved from its initial position. % at some point.

% Even though we also managed to learn a formula that perfectly discriminates agent SGL from the others, we failed to learn a formula completely capturing SME's behaviour.

When searching formulas with a single variable, we split the predicates so that the maximum arity of a fluent is 1. Our algorithm outputs the following very simple formula, which was learnt in a few seconds, while completely characterizing the behaviour of agent ALL:
\begin{align*}
    &\forall x \in \text{Spanner}.\, \lozenge \, \text{carrying}_2(x) \label{lf:spanner21}
    % &\forall x \in \text{Spanner}.\, \text{useable}(x) \, \textsf{U} \, \text{carrying}_2(x) \label{lf:spanner22}
\end{align*}

\section{Conclusion}

In this paper, we have presented a method to learn temporal logic formulas that recognize agents based on examples of their behaviours. We showed that such formulas can be learned using an algorithm that boils down to a reduction to MaxSAT, and that very few examples are sometimes enough to perfectly capture the behaviour of an agent on instances that can differ from the ones used in the training set. This justifies the cost of resorting to a first-order language, which generalizes to new instances, but is also very concise and easily readable by a human. The formulas that we learn can serve as higher-order descriptions of the behaviour of a planning agent.

% \arnaud{Discussion about the fact that the choices that we made ensured some sort of diversity in the formulas learnt}

% \arnaud{Discussion: we often have semantically redundant formulas in the output (ex. $\neg \lozenge$ and $\square \neg$). They can be barred in the MaxSAT encoding too, but we leave this for future work}

% Some of the formulas that we learn are semantically redundant (ex. when the sequence of symbols $\neg \lozenge$ is replaced by $\square \neg$). Even though this is greatly mitigated by our topological guiding and the quantifiers and type constraints we impose (which ensure some diversity in the formulas we learn), we wish to address this issue in future works.

In future works too, we wish to tailor our algorithm and our datasets so that they can generate domain-specific control knowledge. More specifically, we wish to work on its integration into various systems that can be guided with temporal logic, be them automated planning systems~\cite{bacchus00using} or reinforcement learning agents~\cite{icarte2018teaching}. This is in line with previous works on generalized planning, which learn logic-based policies fully capable of solving a set of planning problems, out of a set of example instances and plans~\cite{bonet2019learning,frances2021learning}.

%Some other authors~\cite{bacchus00using} have expressed search control knowledge in a language similar to ours, with the aim of guiding the search of a planner designed to use such knowledge. While this knowledge must be written by a human operator, previous works show that it could also be generated automatically~\cite{delarosa2011learning}. 

%%%%%%%%%%%%%%%%%%%%%%%%%%%%%%%%%%%%%%%%%%%%%%%%%%%%%%%%%%%%%%%%%%%%%%%%

%%% Use this environment to include acknowledgements (optional).
%%% This will be omitted in doubleblind mode.

\begin{ack}
The author would like to thank Martin$\,$C. Cooper for his insightful advices and his careful proofreading of this manuscript. The author also wishes to thank the anonymous reviewers of ECAI 2024 for their remarks and suggestions.
\end{ack}

%%%%%%%%%%%%%%%%%%%%%%%%%%%%%%%%%%%%%%%%%%%%%%%%%%%%%%%%%%%%%%%%%%%%%%%%

%%% Use this command to include your bibliography file.

\bibliography{biblio}

\end{document}